\documentclass[journal,comsoc]{IEEEtran}
\usepackage[T1]{fontenc}
\ifCLASSINFOpdf

\else

\fi

\usepackage{enumerate}
\usepackage{graphicx}
\usepackage{amssymb}
\usepackage{amsmath}
\usepackage{hyperref}
\hypersetup{hypertex=true,
            colorlinks=true,
            linkcolor=blue,
            anchorcolor=blue,
            citecolor=blue}

\begin{document}
	
	\title{An Efficient Style Virtual Try on Network for Clothing Business Industry}
	\author{Shanchen Pang, Senior Member, IEEE, Xixi Tao, Student Member, IEEE, Neal N. Xiong, Senior Member, IEEE, Yukun Dong, Senior Member, IEEE}
	\maketitle

	\begin{abstract}
		With the increasing development of garment manufacturing industry, the method of combining neural network with industry to reduce product redundancy has been paid more and more attention. In order to reduce garment redundancy and achieve personalized customization, more researchers have appeared in the field of virtual trying on. They try to transfer the target clothing to the reference figure, and then stylize the clothes to meet users' requirements for fashion.But the biggest problem of virtual try on is that the shape and motion blocking distort the clothes, causing the patterns and texture on the clothes to be impossible to restore. This paper proposed a new stylized virtual try on network (Style-VTON), which can not only retain the authenticity of clothing texture and pattern, but also obtain the undifferentiated stylized try on. The network is divided into three sub-networks, the first is the user image, the front of the target clothing image, the semantic segmentation image and the posture heat map to generate a more detailed human parsing map. Second, UV position map and dense correspondence are used to map patterns and textures to the deformed silhouettes in real time, so that they can be retained in real time, and the rationality of spatial structure can be guaranteed on the basis of improving the authenticity of images. Third: Stylize and adjust the generated virtual try on image. Through the most subtle changes, users can choose the texture, color and style of clothing to improve the user's experience. As shown in Fig . \ref{fig:fig-01}.Style-VTON use pix22Dsurf method makes virtual try on more generality, in IS and SSIM two evaluation index has achieved the good effect of 2.887 and 0.859, respectively, make the texture and pattern can be efficiently transferred to the user's body, this is very important in the field of electrical business clothing try on, for subsequent can provide users with the most satisfactory clothes, reduce industrial manufacturing production is prepared.
	\end{abstract}
	

	\IEEEpeerreviewmaketitle
	
	\section{Introduction}
	
		\IEEEPARstart{T}{raditional}  virtual try on is based on 3D matching, which is similar to beauty effects. The makeup effects are all virtual images of standard human faces, and the relative position information of facial features is not learned. And the virtual outfit change based on image makes use of generation adversarial network to make the generated pictures close to the real results of the fitting. VITON \cite{1}provides a framework that is widely used, and most current approaches follow a similar framework.
		
		The beginning is the reconstruction learning method using the network generalization ability. Since it is difficult to obtain data sets of the same model wearing different clothes in the same pose, it is common practice to attenuate the supervisory information of the person image and then put the same sample image (the front of the clothes) on top of the person's processed expression. Weak the image's monitoring information to prevent over-referencing and prevent the network from generalizing to different clothes. Secondly, the key points are extracted and the shape is blurred to represent a figure, and the original image is reconstructed. Classical expressions, regardless of clothing or person, use key points of posture, obscure body shapes and head information as input to the network. VITON \cite{1} and CP-VTON \cite{2} both adopted this method. However, the disadvantage of this method is that the generation details are lost. 
		
		Problems with the existing approach include:
		
		\begin{itemize}
			\item The traditional method of texture processing on clothes is not efficient enough to retain extremely complex texture patterns.
			\item The occlusion of the body will have a serious impact on the details of the upper and lower fitting junctions and the short and short payment junctions, and the treatment effect is not ideal.
			\item The clothing style is ordinary, the personal style cannot be reflected, and the user's virtual trying on experience is poor.
		\end{itemize}
	
		In general, they \cite{1,2,3,4} do not guarantee that the texture of the target clothing is well preserved in complex character poses. This brings great hidden trouble and resistance to the realization of virtual outfit system facing real scene application, after all, the user's posture is various when using.
		
		The main contributions of this paper are as follows:(1) UV mapping is used to complete the one-to-one correspondence between the pixel points on the garment and the position pixel points on the user's body, and efficient texture mapping is realized.(2)Fine-grained semantic segmentation preserves the head and face, while better reflecting the details that need to be restored at the intersecting parts of the limbs in the image. 
		
		\begin{figure*}[h]
			\centering
			\includegraphics[width=0.7\linewidth]{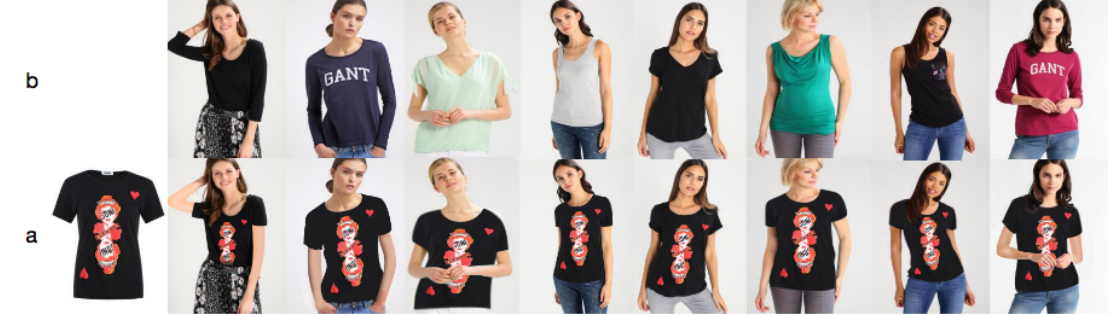}
			\caption{a: the target clothes, b: the reference image of the character, 2 to 8 columns of the fitting result image generated by a and b.}
			\label{fig:fig-01}
		\end{figure*}
		\begin{figure*}[h]
			\centering
			\includegraphics[width=0.7\linewidth]{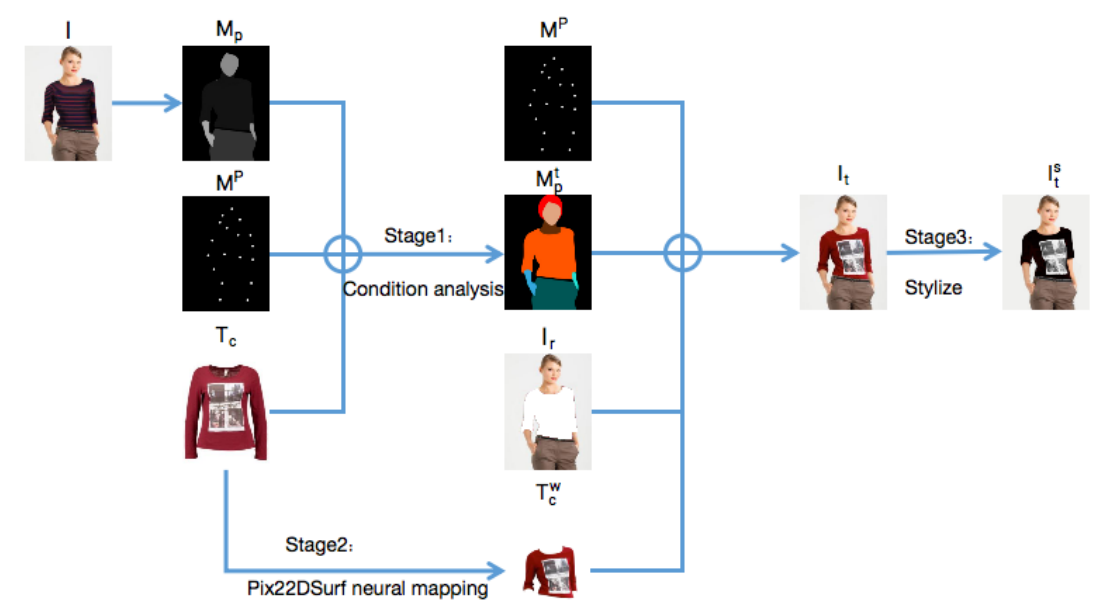}
			\caption{Flowchart of Style-VTON model testing. The first stage: This paper takes the fuzzy segmentation image $ M^p $, the clothing front image $ T_c $ and the target posture heat image $ M_p $ obtained from the user image as the input of the conditional analysis network to predict the body analysis image $ M^t_p $ . The second stage: Pix22DSurf neural mapping was carried out for the front view of the target clothes. Under the guidance of body analysis image $ M^t_p $ and posture heat image $ M_p $ , the user image of the removed clothes and the distorted target clothes were synthesized to generate a realistic image of the user's trying on results $ I_t $. The third stage: the stylized editing network is used to minimize the editing of the fitting results, so as to make the results more fashionable and wearable.}
			\label{fig:fig-02}
		\end{figure*}
	
		The modeling problem of human body and clothing which cannot be covered by existing methods is solved, and artifacts in the generated results are greatly reduced. (3) The user experience has been improved by leaps and bounds in this paper. Just like the average user trying on clothes, they will adjust the way they dress according to their body shape. For example, try on different colors, different styles of the same color and so on. 
		
		This paper finally trained a virtual try on model, which can guide the generation of trying on results of the most suitable style for users, so that users can have more diversity in the choice of clothing collocation. The rest of this paper is organized as follows: Section II.
		
	\section{The Related Work}
	\subsection{Image synthesis}
	Image synthesis is a sample image by the way of segmentation mask to generate the image. The style of the image generation process comes from this sample image. In fact, it allows you to edit any image if you have any image split. By splitting the graph and the sample graph, the algorithms can extract features into the same implicit space using different encoders. They then look for how to distort each other and the sample image, and then add a generator that gets the features from Adain to further improve performance.GAN \cite{10} is one of the most popular generation models in image synthesis, which shows good results in image generation \cite{19,20} and image editing \cite{21,22}. In recent years, there are more researches on image to image conversion around CGN (conditional general adverse network) \cite{23}, which can be applied to different tasks.Converts a given input image into another image with different representation. CGAN shows a powerful effect in image to image task, which aims to transform the input image from one domain to another \cite{24,25,26,27,28,29,30}. For example, an RGB image is generated by its corresponding semantic tag graph. In the image conversion task, these methods have some problems in dealing with the large-scale deformation between the condition image and the target image, such as unreasonable image semantics and unsatisfactory image generation effect. Most misaligned images are used to pan images \cite{1,31,32}, and the quality of the final result is improved by rendering changes from coarse to fine. In the field of virtual trying on, VITON \cite{1} calculates the shape difference between the mask of the front image of the garment and the mask predicted by the human body to ensure a high degree of coincidence during the fitting, and uses thin plate spline interpolation transform (TPS) \cite{33} to maintain the rationality of the garment texture and pattern deformation. But this method is time-consuming in matching two shapes. In addition, the calculated TPS transform is also affected by the prediction mask. Therefore, we propose an efficient texture transmission model, which focuses on the preservation of pattern and texture, and produces no difference fitting results, so that mask matching is efficient and reasonable, and a new coarse to fine strategy is adopted to solve this problem.
	
	\subsection{Virtual try on}
	Virtual trying on is one of the most challenging tasks. Even before the Renaissance of deep learning, virtual try on was an attractive topic. The purpose of the virtual trying on system is that users can choose the clothing collocation through the clothing materials prepared in advance by the software. After selecting the clothes they are interested in, they can take the user's front picture through the camera in the designated shooting area displayed on the screen. Through the photo synthesis after shooting, users can complete the fitting experience, because the models and clothes are real, So the overall image can show the style and texture of the clothes to the maximum extent.In recent years, with the development of deep neural networks, the development of virtual try on has attracted more and more attention. Virtual try-on in deep learning can be divided into two directions, the try-on image generation based on fixed figure pictures and the designated try-on based on 3D mannequin. DRAPE \cite{34} to simulate 2D clothing designs on 3D bodies in different shapes and poses. Most existing virtual try on methods focus on maintaining posture and features. VITON \cite{1}, CP-VTON \cite{2} and other methods adopt the coarse human body shape and take posture mapping as input to generate the try on diagram. VTNFP \cite{4} and other methods use semantic segmentation as input to synthesize the fitting diagram. The method based on UV map will complete the corresponding pixel, and then map the distorted clothes to the corresponding coordinates of the human body. This method does not require the support of a large number of training samples. Efficient to achieve more accurate alignment fitting results.
	
	\subsection{Texture mapping}
	Texture mapping is an important part of realistic image making. It is convenient to make realistic images without spending too much time to consider the surface details of objects. However, the process of texture loading can affect the speed of the program, especially when the texture image is very large. How to properly manage texture and reduce unnecessary overhead is a problem that must be considered in system optimization.A 2D texture mapping is a mapping from a 2D texture plane to a 3D object surface. In general, a 2D texture plane is limited in scope. In this plane region, every point can be expressed by a mathematical function.In this paper, UV mapping is used to solve the problem of low efficiency and low quality in texture mapping." UV" is short for U, V texture map coordinates. It defines information about the location of each point on the image. UV is to map every point on the image to the surface of the 3D model object. The gap position between the points is processed by the software for smooth interpolation, which is called UV.Replacing traditional projection coordinates with UV coordinates, it solves the problem that uneven surface or curved edges in plane projection will result in unsatisfactory joints and deformation as shown in the figure. This usually results in unsatisfactory results, so a lot of tedious work is required before mapping. So, put it on the floor plan where each point on the image corresponds exactly to the surface of the figure. This approach results in better detail and is less likely to cause texture distortion due to excessive distortion in TPS.
	
	\subsection{Maximize activation}
	GAN as a new method for learning generation models, have recently shown good results in a variety of tasks, such as realistic image generation, image processing and text generation. Despite the great success, current GAN models still struggle to produce convincing samples even for image generation with low resolution (such as CIFAR-10) when trained on data sets with high variability. At the same time, it has been found by experience that the use of category labels can significantly improve the quality of samples.Activation maximization \cite{35} is a gradient based method to optimize images and highly activate target neurons in the neural network. It is often used to visualize the neurons of a pre-trained neural network. GAN training can be regarded as the process of antagonistic activation maximization. Specifically, the training generator maximizes the activation of each generated sample in the neuron representing the target log-like probability, while training the discriminator to distinguish the generated samples and prevent them from getting higher activation. It is worth mentioning that the sample that maximizes the activation of a neuron is not necessarily of high quality, and various priors have been introduced to combat this phenomenon. In GAN, the confrontation process of GAN training can detect unrealistic samples to ensure that they come from high-quality samples, which will also strongly confuse discriminators. It is widely used for visualizing what a network has learned \cite{35,36,37,38, 39}, and recently to synthesize images \cite{40,41}.In particular, \cite{40} also generates clothing images, but they generate single-garment products rather than full body out-fits. 
	
	\subsection{Geometric deformation}
	At present, 2D virtual trying on display is mostly realized by image processing technology. In 2D virtual try on, based on the fitting effect of personalized shape simulation become the research focus, namely how to implement the shape parameter customization lets users according to their corresponding clothing models, and clothing models based on shape feature deformation by reasoning according to certain rules of garment processing, thus complete personalized virtual controls and try it on. In the process of 2D virtual trying on based on personalized body shape, clothing image deformation technology is also a key problem. The effect of clothing image deformation directly affects the overall satisfaction of online shopping.In the generation model, the problem of large space deformation is mainly studied in the background of the generation of human image guided by attitude. This task involves generating an image of a person, given a reference person and a target pose. Some approaches use unentanglement to separate posture and shape information from appearance, allowing for the reconstruction of reference persons for different poses \cite{42,43,44}. However, the most recent generation methods of pose guides involve explicit spatial transformations in their buildings, whether or not they are learned \cite{45}. Through partial specific learnable affine transformation, different body parts of a person are segmented and moved to the target posture, and applied to the pixel level. The deformable GAN from \cite{46} is a U-Net generator whose jump connections are deformed by a partially specific affine transformation. These transformations are calculated based on the attitude information of the source and target. Instead, \cite{3} uses a convolutional geometry matcher from 
	to learn the thin plate spline (TPS) transformation between source artificial parsing and synthesis target parsing and alines the depth feature mapping of the encoder decoder generator.
	
	\subsection{Fashionality and compatibility}
	Fashion has been studied extensively because of its huge profit potential. Existing methods mainly focus on clothing analysis, clothing identification through attributes, providing product styles and matching with clothing seen on the street or connected with factories,fashion recommendation,visual compatibility learning, and fashion trend prediction. In contrast to these efforts, our approach focuses on virtual try on, using only 2D images as input, and using an encoder to control fashion attributes. In contrast to more challenging tasks, the most recent work requires only modifying properties (for example, colors and textures).
	
	Visual compatibility plays an essential role in fashion recommendation. Metric learning based methods have been adopted to solve this problem by projecting two compatible fashion items close to each other in a style space. Unlike these approaches which attempt to estimate fashion compatibility,this paper incorporate compatibility information into an image in painting framework that generates a fashion image containing complementary garments. Furthermore, most existing systems rely heavily on manual labels for supervised learning. In contrast, we train our networks in a selfsupervised manner, assuming that multiple fashion items in an outfit presented in the original catalog image are compatible to each other, since such catalogs are usually designed carefully by fashion experts. Thus, minimizing a reconstruction loss jointly during learning to in paint can learn to generate compatible fashion items.

	\section{Our PROOSED  STYLE-VTON}
		In this paper, an efficient mapping model Pix22Dsurf was designed to process 2D image texture details, making significant progress in the fitting results of texture, shape and other attributes. On the trying on effect, the editor with diversified design styles makes the fitting result closer to the aesthetic needs of users. Inspired by multi-pose guidance \cite{3} and VITON \cite{1}, image processing is divided into three stages. The first stage: process the reference image of the character, analyze the costume picture, retain the irrelevant body parts, obtain the posture heat image with posture estimation \cite{5, 6}, and generate the character analysis picture with DenseNet \cite{7}. The second stage: 2D texture mapping, learning an efficient neural mapping and transmitting the texture in real time. The third stage: make changes to the existing clothes at the minimum to ensure that the original appearance and style of clothes will not be changed under the premise of increasing fashion degree \cite{8, 9}. The generated pictures successfully tried on were stylized and adjusted to improve the user's satisfaction, as shown in Fig.\ref{fig:fig-02}.
		
		\subsection{Condition Analysis Network}
			The main challenge of virtual try on is to transform the target image into an image suitable for the body posture, while retaining the texture. In order to keep the expression on the body and clothes from being associated without some details,we need to complete the parsing of the body, we use target clothing images, body posture key points and fusion of the semantic template of the human body as an CGAN \cite{10, 11} to input, generate human parsing images. Since the change of human posture will lead to the deformation of clothing to different degrees, this paper uses the posture estimator \cite{5, 6} to extract the features of posture information. The human body parser \cite{12,13,14} is used to calculate a human body segmentation map and extract the binary mask. Unrelated information, such as facial information and hair parts, is preserved. In this way, the pose estimator and body parser can effectively guide the synthesis of precise areas of body parts.
			
			\begin{figure*}[h]
				\centering
				\includegraphics[width=1\linewidth]{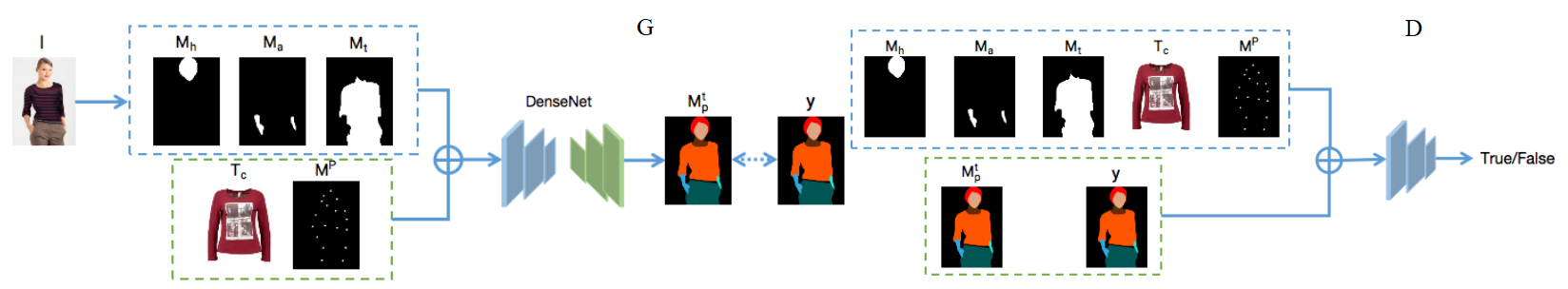}
				\caption{The first-phase network architecture of Style-VTON. Generator: Conditional analytic learning module generates predicted character analytic images from a network of garment-guided postures. Discriminator: The game is helpful to generate high quality body analysis. }
				\label{fig:fig-03}
			\end{figure*}

			In general, the predictive body parsing image is learned given the user image, the front of the cloth image, and the target posture heat image. Firstly, the body parser is used to extract the head mask, arm mask and drive mask, which are fused into an indistinguishable region to obtain the input item. Posture estimator computes heat images of key postural points. Combined with the front image of clothing as the input of the conditional analytic network, the human body image of the target posture is generated through DenseNet.Conditional analytic network is shown in Fig. \ref{fig:fig-03}.
			
			The loss function of the conditional analytic network is similar to that of CGAN, as shown in equation (\ref{equa-01}) and the ultimate minimum and maximum loss is:
			\begin{equation}\label{equa-01}
				\begin{aligned}
					L_{adv1} = min_G max_D V(G, D)\\
					=\mathbb{E}_{M_p, T_c, I\thicksim P_{data}}\left[ \log\left(1 - D\left( G\left( M_p, T_c, I \right) \right) \right) \right]\\
					+ \mathbb{E}_{y, M_p, T_c, I\thicksim P_{data}}\left[ \log D\left( y, M_p, I, T_c \right) \right].
				\end{aligned}
			\end{equation}
		
			Calculate the per-pixel loss of the generated human body analytic image as shown in equation (\ref{equa-02}):
			\begin{equation}\label{equa-02}
				L_{parsing} = \mathbb{E}_{y, M_p, T, I\thicksim P_{data}},
			\end{equation}
			
			Where, $ P_{data} $ represents the distribution of real data, $ M_p $ represents the input, $ T_c $ is the head mask, arm mask and drip-dry mask splicing,  is the front image of the target clothes, $ I $ is the human figure, and  is the real human body analytic mask. $ L_1 $ is a counter loss, $ L_{parsing} $ representing the pixel level loss.
			
		\subsection{Pix22DSurf Neural Mapping}
			Inspired by the achievements of UV map in the field of face reconstruction, the main idea of this paper is to learn the UV mapping from image to clothing. Instead of using texture information, only using silhouette shapes can automatically generate the texture corresponding to the distorted clothing.
			
			The purpose of traditional UV map is to map the 2D texture to the 3D model, which provides a connection between the 2D human surface and the texture image. Because the images on the Internet are different in texture,posture and background, it is difficult to achieve accurate texture conversion. This paper aims to complete the dense correspondence of 2D without losing the texture.  
			
			The clothing mask after obtaining contour information is treated as UV map to store all pixel information, and the clothing contour is transferred according to the body surface. Although the body surface is 2D, it can still be covered with 3D UV map. In this paper, a CNN with clothing mask as input is trained to predict the corresponding relationship from UV image of clothing surface to image pixel position. The texture coordinates of the object object that is distorted in this way do not change with the vertex of the template mesh. After interpolation, the texture of the object object also changes as if it were distorted in the template mesh immediately following it. This sub-model learns the inherent differences in the relative shape and posture of the appearance, and is therefore summarized as a variety of clothing images with different textures. Learned an efficient neural mapping pix22Dsurf, custom non-rigid pixel to two-dimensional surface pixel registration method, automatic calculation of the alignment of 2D image pairs, used for real-time image texture transfer to the grid.
			
\begin{figure}[h]
	\centering
	\includegraphics[width=1\linewidth]{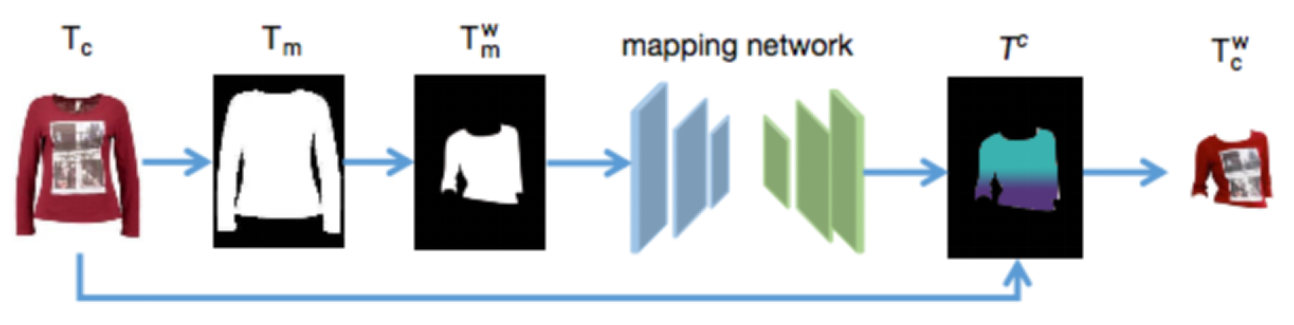}
	\caption{Network architecture in the second stage of Style-Viton: The mask with the target contour information is used as the input of network mapping for Pix22Dsurf to map the texture to the surface of the UV image to generate the target clothing mapped by the texture $ T_c^w $.}
	\label{fig:fig-04}
\end{figure}
		
			The idea of this paper is to learn the dense correspondence of a 2D UV map from the surface of the clothing to the surface of a 2D character's body surf using only the contour information, as shown in Fig. \ref{fig:fig-04}. Firstly, the contorted clothing contour information should be extracted, and the clothing mask $ T_m $ and posture heat image $ M^p $ obtained by the target clothing $ T_c $ should be used as input to generate the contorted mask with the user's body shape information (the contorted), as shown in Fig. \ref{fig:fig-05}. Second, following a geometric approach, learning the mapping  from the mask  (which has only contour information) to the UV corresponding mapping  forces the network to infer the shape of the input. This will effectively learn to predict the corresponding relationship  between the coordinates of the pixel points  on each UV map position point  and the corresponding image, and minimize the following losses during the training.as shown in equation (\ref{equa-03}):
			
			\begin{figure*}[h]
				\centering
				\includegraphics[width=1\linewidth]{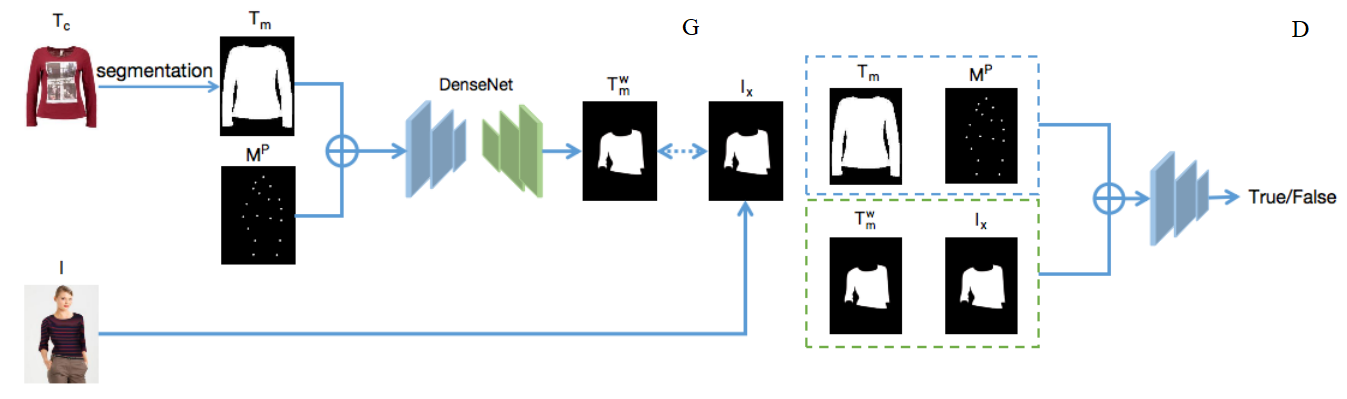}
				\caption{Training diagram for obtaining target clothing profile. Similar to the method in Fig. \ref{fig:fig-03}, the generator and discriminator are pitted against each other to produce a target contour mask with accurate results.}
				\label{fig:fig-05}
			\end{figure*}

			\begin{equation}\label{equa-03}
				L_{total} = \lambda_1 L_{adb} + \lambda_2 L_1 + \lambda_3 L_{recon},
			\end{equation}
			
			Where in $ L_{adv} $ is adversarial loss as shown in equation (\ref{equa-04})
			
			\begin{equation}\label{equa-04}
				\begin{aligned}
						L_{total} = \min_G \max_D V (G, D) \\
						= \mathbb{E}_{M_p}\left[ \log\left( 1 - D\left( G(z) \right) \right) \right] .
				\end{aligned}
			\end{equation}
			
			$ L_1 $ is the regularization loss of L1as shown in equation (\ref{equa-05})
			
			\begin{equation}\label{equa-05}
				L_1 = || I_X - T^w_c ||_1 .
			\end{equation}
			
			In the regularization loss of L1, $ I_x $is the silhouette of real human clothing after projection,$ T^w_c $is a distorted mask.
			
			$ L_{recon} $is the reconstruction loss, is the loss between the predicted texture map obtained by mask and the target texture map by minimized as shown in equation (\ref{equa-06}):(the texture obtained by network prediction mapping relationship and the texture obtained by cut)
			
			\begin{equation}\label{equa-06}
				 L_{recon} = \sum_{i}^{N} \sum_{k, l}^{K, L} \Arrowvert \left[f^{1}_{k, l} ( T_{m}^{w^i}; w ), f^2_{k, l}(T_{m}^{w^i}; w)\right] - T_{c_{k, l}}^{w^i}  \Arrowvert_1
			\end{equation}
			
			In the reconstructed loss $ f(T_{m}^{w^i}; w) $ is the predictive texture mapping relationship and $ T_{c_{k, l}}^{w^i} $ is the target texture map.
			
			The results of stage 1 and 2 are guided to generate the try-on result image. During the training, the counter loss should be minimized, as shown in equation (\ref{equa-07}):
			
			\begin{equation}\label{equa-07}
			\begin{aligned}
& L_{adv3} = \min_G \max_D V(G, D)\\
& =\mathbb{E}_{M_p, M^t_p, T^w_c, I_r \sim P_{data}} \Bigl[ \log\Bigl( -D\Bigl( G\bigl( M_p, M^t_p, T^w_c, I_r \bigr), \\
& \qquad\qquad\qquad\qquad\qquad\qquad\qquad M_p, M^t_p, T^w_c, I_r  \Bigr) \Bigr)  \Bigr]\\
& + \mathbb{E}_{I_{GT}, M_p, M^t_p, T^w_c, I_r \sim P_{data}}[ \log D( I_{GT}, M_p, M^t_p, T^w_c, I_r ) ] . 
			\end{aligned}
			\end{equation}
		\subsection{Style Generation Network}
			In order to control the idealization of style, change the styles of shirts and t-shirts in the aspects of neckline and cuff, and identity or other factors unrelated to fashion can be kept unchanged at the same time. Therefore, we modeled their spatial position and needed to control the local texture and shape at the same time. Because the normal try on process is based on the selection of styles, different styles can be adjusted for different needs, such as longer or shorter sleeves, shorter waist or looser. Considering many factors affecting fashion, this paper designs a style generation network, which can control the design and trying on.
			
			Based on depth image characteristics of the local code, the character images in the controls (such as t-shirts, pants, accessories) mapped to the respective code, by editing module, maximum use of neuron activation method step by step to update the control code, maximize clothing score, to generate the same user experience of stylized adjusted clothing.
			
\begin{figure*}[h]
	\centering
	\includegraphics[width=1\linewidth]{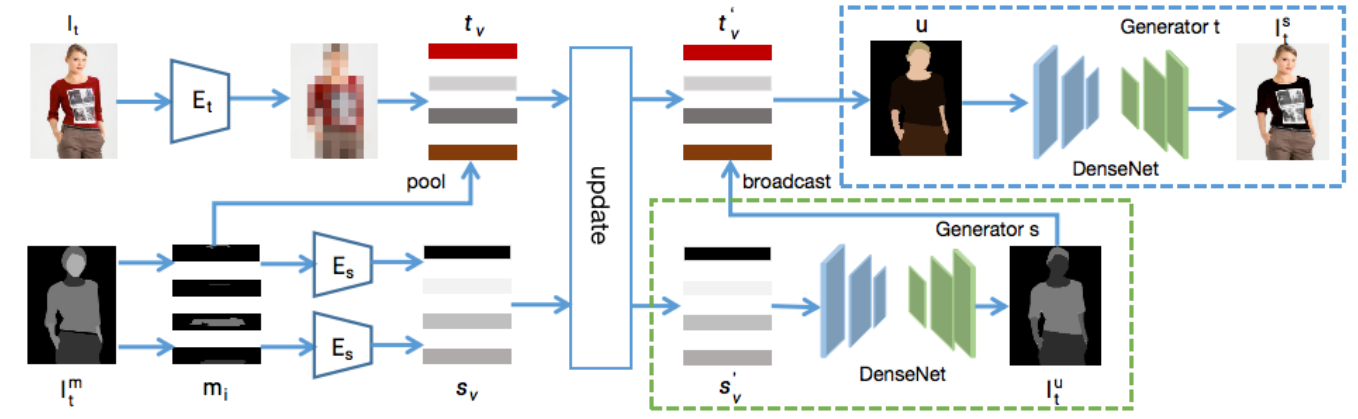}
	\caption{Stylized generated network architecture. The texture features $s_v$ and shape features $ t_v $ are obtained from texture encoders $ E_t $ and shape encoders$ E_s $ with the results of the first and second stages and their gray-scale semantic segmentation images as input. The update module will edit and update the feature vectors. After an edit, the shape generator $ G_s $ will decode the updated shape features to generate a new semantic segmentation gray-scale image $ I^u_t $ , which broadcasts the texture features $ s_v $ to generate input from the texture generator $ G_t $ with the shape features, and finally generates the style-adjusted target image $ I^s_t $.}
	\label{fig:fig-06}
\end{figure*}

			Our editor edits textures and shapes separately to produce stylized images. (1) Texture features. The input image $ I_t $ is a real image of a person wearing clothes. Area maps $ m^i \in m $ assign each pixel to an area of a fixed garment or body part. Since this paper only adjusts the clothing style of the upper body, a unique regional label defined in Chictopia10k is adopted: face, hair, hat, T-shirt, etc. First, input $ I_t $ into a learned texture encoder $ E_t $ , get the feature vector, and average pool the texture of the specified label area, that is, get the texture feature vector of the area $ m^i $ . (2) Shape features. The shape of each area controls the separation of textures. Specifically, a binary segmentation image is constructed for each region, and a shared shape encoder is used to encode $ E_s $ each region image into a shape feature. As shown in Fig. \ref{fig:fig-06}.
			
			The shape codes designed in this paper are variational auto-encoder (VAE, VAE could learn the probability distribution of clothing shape, assuming that the posterior distribution is $ E_s(z | m) $. During the training, KL divergence is used to measure the difference between distributions, and KL divergence needs to be minimized,as shown in equation (\ref{equa-08}):
			
			\begin{equation}\label{equa-08}
				D_{KL} \left( E_s(z | m) \parallel p(z) \right) , 
			\end{equation}
			
			Where $ p(z) $ is a Gaussian with a mean of 0 and a variance of 1.
			
			Texture encoder is a conditional generation adversarial network,which generates regional texture mapping  based on texture features guided by regional images . Therefore, it is necessary to minimize the adversarial loss of CGAN,as shown in equation (\ref{equa-09}):
			
			\begin{equation}\label{equa-09}
			\begin{aligned}
				L &= \min_{G^t, E^t}\max_D W(G_t, E_t, D)\\
				 &= \mathbb{E}_{(m, x)} \left( \log D(m, I_t) + \log\left( 1 - D(m, G_t(m, u) ) \right) \right) , 
			\end{aligned}
			\end{equation}
			
			Where  is the region diagram, $ I_t $ is the input image, $ G_t(m, u) $ is the generated style after the minimum edit virtual try on image.

	\section{Performance Analysis}
		Due to the lack of data on full-body dressing of real people to solve the problem in this paper, the three stages of style fitting network were trained respectively. The evaluation method is to first compare the visual truthfulness and diversity of images with other methods and discuss the results quantitatively.
		
		\subsection{Data Set}
			The data set used in this article is divided into two parts.
			
			Part 1:Look for a pair of images that include the front of the person and the front of the jacket. VITON is the data set collected by Zalando clothing website. Based on VITON, this paper expands and selects from Zalando, Tom Tailor, Jack and Jones. Containing approximately 17,000 image pairs, the 17,000 image pairs were divided into 14,875 training pairs and 2,125 test pairs. Because it is difficult to obtain full-body images and the resolution is too low, the images collected in this paper are all virtual tries on tops. The tops have certain representativness, diversified patterns and complex postures, which can better express the trying effect.
			
			Part 2:We want to train a fashion classifier to perform minimal editing, and the ideal training set should consist of pairs of pictures, each of which shows the same person wearing slightly different clothes, one of which is considered more stylish than the other. This paper uses the ChicTopia10K data set for experiments. 15,930 images were used to train the generator and 12,744 classifiers. 3,240 of these unfashionable examples were evaluated.
			
		\subsection{Implementation Details}
			The size of all input and output images is fixed at $ 256 \times 192 $.
			
			Training test Settings. This paper put forward by the training module, and put them together, final output stylized try on images, by setting the weight of each training module, on the condition of analytic network, Pix22Dsurf neural mapping (including contour generation and texture mapping phase), stylized generation network VAE, GAN and classifier trained 200,70,100,300,200,120 epoch respectively, super parameter settings , Adam optimizer, fixed learning rate is 0.0002. The step size of the training generator is 15K, the step size of the improved network is 9K, and the size of batchsize is 16. This article was implemented with the deep learning toolkit PyTorch and experimented on the Ubuntu 18.04 platform using two NVIDIA Geforce 3090 GPUs. The testing process is similar to the training process, except that the target clothing is different from the clothing in the reference image. In this paper, the model is tested on the test set and the results are evaluated qualitatively and quantitatively.
			
			Conditional generation network stage. Each generator of Style-VTON likes a DenseNet network from the Dense Block module: including BN, relu, conv($ 3*3 $),and dropout. Transition layer module: consists of BN, relu, conv$ (1*1) $, dropout, pooling($ 2*2 $). The discriminator applies the same architecture as pix2pixHD\cite{15}. Each discriminator contains four lower sample layers, including InstanceNorm and LeakyReLU activation functions.
			
			Pix22Dsurf Neural mapping phase. For split networks, use UNet. For the mapping network, use six blocks ResNet \cite{16}. The choice of regularization and activation functions is consistent with \cite{17}. UV mapping: only the front faces of the characters are learned, RCNN structure with pyramid network (FPN) features is adopted, and ROI align pooling is used to obtain the labels and coordinates of the dense part in the selected area, that is, the upper torso part.
			
			The network of style generation. All the generation networks were trained from zero. For VAE, we maintained the same learning rate of 0.0002 in the first 100 epochs and the linear decay learning rate of 0 in the next 200 epochs. For GAN, the structure adopts \cite{18}. The same learning rate is maintained in the first 100 epochs at 0.0002, and the learning rate is 0 as a result of the linear decay of the next 100 epochs. For the fashion classifier, the weight attenuation is 0.0001 and the learning rate is 0.001.
			
		\subsection{Qualitative Results}
\begin{figure}[h]
	\centering
	\includegraphics[width=1\linewidth]{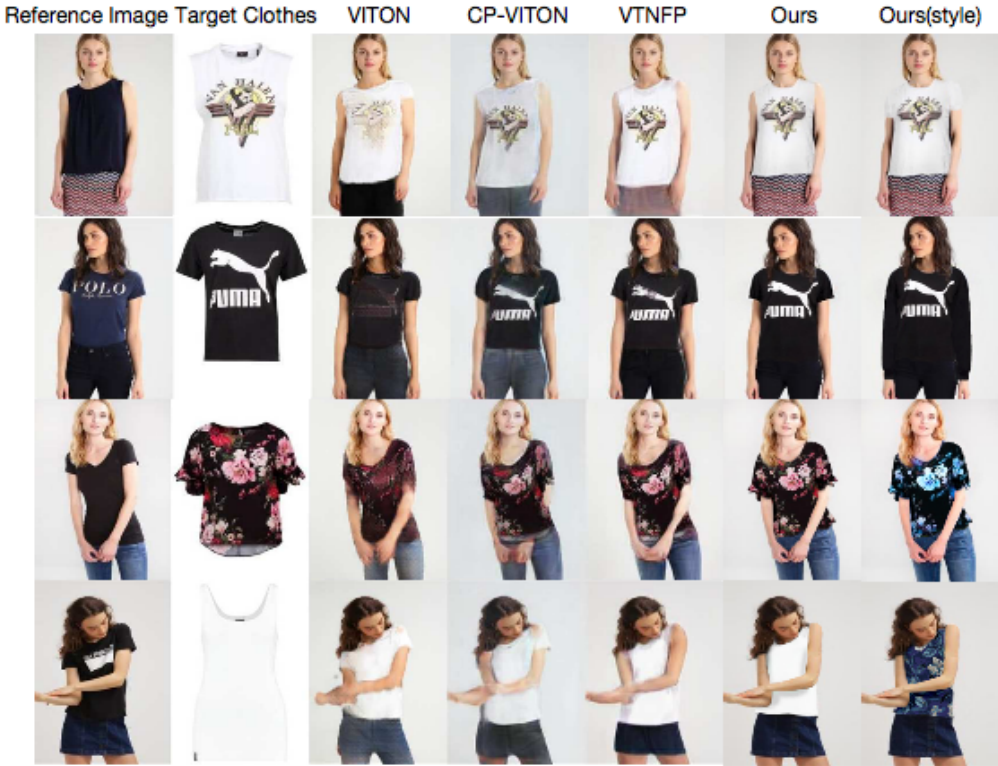}
	\caption{As shown in the figure above, the first column is the reference figure image, the second column is the target clothes, and the third to sixth columns are the fitting results of VITON,CP-VTON,VTNFP and our method respectively. The seventh column is the stylized trying on results.}
	\label{fig:fig-07}
\end{figure}

			In this section, some qualitative results of the model are provided. Visualizations demonstrate the contribution of various network components that we incorporate into the model to the performance of the model. This result indicates that compared with VITON \cite{1}, CP-VTON \cite{2}, VTNFP \cite{4} and ACGPN \cite{6}, Style-VTON produces more realistic virtual try on images. As shown in Fig. \ref{fig:fig-07}.
			
			In the first and second stages of this paper, various problems existing in the above four networks are solved. In the case of clear generated results and no distortion, Pix22Dsurf can increase the processing speed of several orders of magnitude and map the texture. At the same time, in order to improve users' satisfaction with virtual trying on results,in the third stage, the method of this paper makes minimal editing of the generated sample fitting results, so as to better meet the requirements of human body type diversification.
			
		\subsection{Quantitative results}
			In this paper, structural similarity (SSIM) was used to evaluate the similarity between the composite image and the real image, to verify the performance of image generation, and IS to evaluate the visual quality and diversity of the composite image. The higher the score on these two indicators, the higher the quality of the results. Table \ref{tab-01} lists the SSIM and IS scores given by Style-VTON. In the stylized fitting stage, this paper evaluates the model performance through the degree of fashion and the amount of change after editing. The measure of fashion improvement is the distance between the original work and the generated result and the human perception of the generated result. In this paper, users of all ages were asked to rate the results in Table \ref{tab-02}.slight fashion improvement/little change.
			
			\begin{table}[!t]
				\renewcommand{\arraystretch}{1.3}
				\caption{The results of compared quantitatively with three method}
				\label{tab-01}
				\centering
				\begin{tabular}{ccc}
					\hline\hline
					Method & IS & SSIM\\
					\hline
					VITON & 2.650 & 0.783\\
					VTNFP & 2.784 & 0.803\\
					CP-VTON & 2.757 & 0.745\\
					Ours(Style-VTON) & 2.887 & 0.859\\
					\hline
				\end{tabular}
			\end{table}
			
			In order to further evaluate the performance of our model, user perception research was conducted. In this study, we designed an A/B test to compare the quality of images generated by VITON, VTNFP, CP-VTON and Style-VTON. We recruited 80 volunteers and showed them 500 sets of test data, each consisting of four images: the generated result, the target clothing, the parsed body image and the distorted clothing image. Each volunteer was randomly assigned 50 sets of test data from two methods and asked to choose from each set the composite image that he or she thought was of better quality.
			
			In the A/B test of Style-VTON and VITON, 77.87\% of the images generated by Style-VTON were selected by the volunteers to be of good quality. A/B test was conducted between Style-VTON and CP-VTON, and 81.38\% of the volunteers chose the images generated by the former. These random tests confirm the qualitative results shown in the previous section, which prove that Style-VTON performs significantly better than the previous model.
			
			\begin{table}[!t]
				\renewcommand{\arraystretch}{1.3}
				\caption{A/B test results: Our method is compared with the other three methods to select the ratio of image authenticity}
				\label{tab-02}
				\centering
				\begin{tabular}{@{ }c@{ }c@{ }c@{ }c@{ }c@{ }c@{ }}
					\hline\hline
					Method & Proportion & Method & Proportion & Method & Proportion \\
					\hline
					VITON & 22.13\% & VTNFP & 18.62\% & CP-VTON & 33.24\% \\
					Ours & 77.87\% & Ours & 81.38\% & Ours & 66.76\% \\
					\hline
				\end{tabular}
			\end{table}
		
		\subsection{Engineering applications}
			Virtual trying on makes use of the Internet and AI as a driving force to boost the upgrading of the textile and apparel industry and improve the shopping experience of consumers. Consumers can browse a variety of exquisite clothes with virtual try on technology and then try on them through the virtual try on room. Consumers with their real figure scale and characteristics, experience immersive virtual try on, you can use the dressing interactive experience in-depth and personalized virtual, to determine whether the clothes fit, the fitting saves time and effort, with regard to business, and also reduce the artificial service cost, increase the volume, at the same time, the fitting experience full of fun, It also brings more customers to businesses.
			
\begin{figure}[h]
				\centering
				\includegraphics[width=0.5\linewidth]{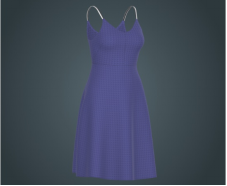}
				\caption{As shown in the figure above,the digital modeling of garments contributes to the industry and users' understanding of design and fabrics.}
				\label{fig:fig-08}
			\end{figure}

			Virtual try on with efficient texture mapping preserves the texture and pattern of the garment,allowing the fabric ready-to-wear effect to be previewed online,as shown in Fig. \ref{fig:fig-08}.
			
			In addition to let consumer experience a virtual try on finished garments, virtual trying on technology can also be used in the industries of fabrics, make fabric seconds "garment", to the effect of the garment to show to the customers, to help designers and fabric garment fabrics purchasing online preview effect, improve the work efficiency, to quickly determine the fabric and pattern for attaining costume design is in line with the expected effect, Save time and cost.Virtual trying on carries and stores complete fabric information, such as fabric attribute information and picture information, which can realize the delivery of realistic fabric clothing effect, help fabric merchants to display fabric goods more intuitively, provide a brand new merchant interaction mode in the industry, and unlock new technology in the digital industry of the textile industry.
			
\begin{figure}[h]
	\centering
	\includegraphics[width=1\linewidth]{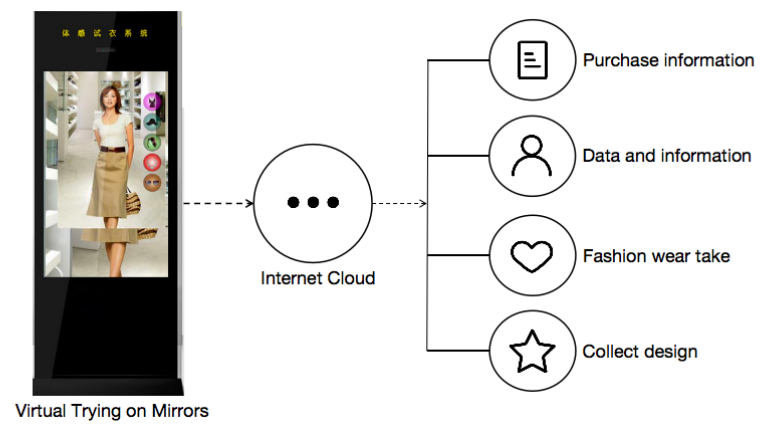}
	\caption{The development trend of virtual trying on in clothing customization industry}
	\label{fig:fig-09}
\end{figure}

			At the same time, the virtual try on system will be mainly used for finished products and personalized products of the e-commerce business, through the Internet, in the case of not easy to contact the physical simulation of the final effect using technical means. In commercial physical stores, the use of virtual trying on technology can greatly improve the store stopping rate, transaction rate and reputation. In the creative stage of the product, customers can experience the final effect, which eliminates the process of making samples, greatly improves the customization efficiency and customer experience, and also provides a technical basis for remote personalized customization based on the Internet. It will not only provide good experience for customers, but also greatly improve the sales and service mode, and eventually become the front-end entrance of product customization,as shown in Fig. \ref{fig:fig-09}.
			
	\section{The Conclusion and Future Work}
			
		This paper proposes a simple and effective virtual try on model based on images, using a three-stage design strategy, so that the texture in the fitting results can be well mapped in the context of complex figure posture and background. Finally, the resulting fitting results are stylized and minimized to produce attractive results. This paper introduces the innovation of pix22Dsurf method to improve the detail quality of image synthesis. Both the quantitative evaluation and the user perception opinion prove that the method in this paper is able to produce better results.
		
		Discussion: work in the future, plans to expand the source of training data, for example, you can use a wider range of social media platforms such as weibo, sets, twitter, seize the social classification of wind direction to vogue, for thinning controls, to generate the style of personal preference or special occasions dress is recommended, which promote the demand of the people, the custom apparel industry, reduce the redundancy of the garment industry in the industry, reduce waste.

	\bibliography{IEEEabrv,ref}

\begin{thebibliography}{10}
\providecommand{\url}[1]{#1}
\csname url@samestyle\endcsname
\providecommand{\newblock}{\relax}
\providecommand{\bibinfo}[2]{#2}
\providecommand{\BIBentrySTDinterwordspacing}{\spaceskip=0pt\relax}
\providecommand{\BIBentryALTinterwordstretchfactor}{4}
\providecommand{\BIBentryALTinterwordspacing}{\spaceskip=\fontdimen2\font plus
\BIBentryALTinterwordstretchfactor\fontdimen3\font minus
  \fontdimen4\font\relax}
\providecommand{\BIBforeignlanguage}[2]{{%
\expandafter\ifx\csname l@#1\endcsname\relax
\typeout{** WARNING: IEEEtran.bst: No hyphenation pattern has been}%
\typeout{** loaded for the language `#1'. Using the pattern for}%
\typeout{** the default language instead.}%
\else
\language=\csname l@#1\endcsname
\fi
#2}}
\providecommand{\BIBdecl}{\relax}
\BIBdecl

\bibitem{1}
X.~Han, Z.~Wu, Z.~Wu, R.~Yu, and L.~S. Davis, ``Viton: An image-based virtual
  try-on network,'' in \emph{Proceedings of the IEEE Conference on Computer
  Vision and Pattern Recognition (CVPR)}, June 2018.

\bibitem{2}
B.~Wang, H.~Zheng, X.~Liang, Y.~Chen, L.~Lin, and M.~Yang, ``Toward
  characteristic-preserving image-based virtual try-on network,'' in
  \emph{Proceedings of the European Conference on Computer Vision (ECCV)},
  September 2018.

\bibitem{3}
H.~Dong, X.~Liang, X.~Shen, B.~Wang, H.~Lai, J.~Zhu, Z.~Hu, and J.~Yin,
  ``Towards multi-pose guided virtual try-on network,'' in \emph{Proceedings of
  the IEEE/CVF International Conference on Computer Vision (ICCV)}, October
  2019.

\bibitem{4}
R.~Yu, X.~Wang, and X.~Xie, ``Vtnfp: An image-based virtual try-on network with
  body and clothing feature preservation,'' in \emph{Proceedings of the
  IEEE/CVF International Conference on Computer Vision (ICCV)}, October 2019.

\bibitem{10}
\BIBentryALTinterwordspacing
I.~J. Goodfellow, J.~Pouget-Abadie, M.~Mirza, B.~Xu, D.~Warde-Farley, S.~Ozair,
  A.~C. Courville, and Y.~Bengio, ``Generative adversarial nets,'' in
  \emph{NIPS}, 2014, pp. 2672--2680. [Online]. Available:
  \url{http://papers.nips.cc/paper/5423-generative-adversarial-nets}
\BIBentrySTDinterwordspacing

\bibitem{19}
\BIBentryALTinterwordspacing
M.~Ranzato, J.~M. Susskind, V.~Mnih, and G.~E. Hinton, ``On deep generative
  models with applications to recognition,'' in \emph{CVPR}, 2011, pp.
  2857--2864. [Online]. Available:
  \url{https://doi.org/10.1109/CVPR.2011.5995710}
\BIBentrySTDinterwordspacing

\bibitem{20}
L.~Zhang, G.-J. Qi, L.~Wang, and J.~Luo, ``Aet vs. aed: Unsupervised
  representation learning by auto-encoding transformations rather than data,''
  in \emph{Proceedings of the IEEE/CVF Conference on Computer Vision and
  Pattern Recognition (CVPR)}, June 2019.

\bibitem{21}
P.~Isola, J.-Y. Zhu, T.~Zhou, and A.~A. Efros, ``Image-to-image translation
  with conditional adversarial networks,'' in \emph{Proceedings of the IEEE
  Conference on Computer Vision and Pattern Recognition (CVPR)}, July 2017.

\bibitem{22}
\BIBentryALTinterwordspacing
Y.~Li, C.~Fang, J.~Yang, Z.~Wang, X.~Lu, and M.~Yang, ``Universal style
  transfer via feature transforms,'' \emph{CoRR}, vol. abs/1705.08086, 2017.
  [Online]. Available: \url{http://arxiv.org/abs/1705.08086}
\BIBentrySTDinterwordspacing

\bibitem{23}
J.~Fajtl, V.~Argyriou, D.~Monekosso, and P.~Remagnino, ``Amnet: Memorability
  estimation with attention,'' in \emph{Proceedings of the IEEE Conference on
  Computer Vision and Pattern Recognition (CVPR)}, June 2018.

\bibitem{24}
\BIBentryALTinterwordspacing
T.~Kim, B.~Kim, M.~Cha, and J.~Kim, ``Unsupervised visual attribute transfer
  with reconfigurable generative adversarial networks,'' \emph{CoRR}, vol.
  abs/1707.09798, 2017. [Online]. Available:
  \url{http://arxiv.org/abs/1707.09798}
\BIBentrySTDinterwordspacing

\bibitem{25}
C.~Ledig, L.~Theis, F.~Huszar, J.~Caballero, A.~Cunningham, A.~Acosta,
  A.~Aitken, A.~Tejani, J.~Totz, Z.~Wang, and W.~Shi, ``Photo-realistic single
  image super-resolution using a generative adversarial network,'' in
  \emph{Proceedings of the IEEE Conference on Computer Vision and Pattern
  Recognition (CVPR)}, July 2017.

\bibitem{26}
\BIBentryALTinterwordspacing
J.~J. Zhao, M.~Mathieu, and Y.~LeCun, ``Energy-based generative adversarial
  network,'' \emph{CoRR}, vol. abs/1609.03126, 2016. [Online]. Available:
  \url{http://arxiv.org/abs/1609.03126}
\BIBentrySTDinterwordspacing

\bibitem{27}
\BIBentryALTinterwordspacing
T.~Wang, M.~Liu, J.~Zhu, G.~Liu, A.~Tao, J.~Kautz, and B.~Catanzaro,
  ``Video-to-video synthesis,'' \emph{CoRR}, vol. abs/1808.06601, 2018.
  [Online]. Available: \url{http://arxiv.org/abs/1808.06601}
\BIBentrySTDinterwordspacing

\bibitem{28}
R.~Li, J.~Pan, Z.~Li, and J.~Tang, ``Single image dehazing via conditional
  generative adversarial network,'' in \emph{Proceedings of the IEEE Conference
  on Computer Vision and Pattern Recognition (CVPR)}, June 2018.

\bibitem{29}
Y.~Choi, M.~Choi, M.~Kim, J.-W. Ha, S.~Kim, and J.~Choo, ``Stargan: Unified
  generative adversarial networks for multi-domain image-to-image
  translation,'' in \emph{Proceedings of the IEEE Conference on Computer Vision
  and Pattern Recognition (CVPR)}, June 2018.

\bibitem{30}
K.~E. Ak, J.~H. Lim, J.~Y. Tham, and A.~A. Kassim, ``Attribute manipulation
  generative adversarial networks for fashion images,'' in \emph{Proceedings of
  the IEEE/CVF International Conference on Computer Vision (ICCV)}, October
  2019.

\bibitem{31}
G.~Balakrishnan, A.~Zhao, A.~V. Dalca, F.~Durand, and J.~Guttag, ``Synthesizing
  images of humans in unseen poses,'' in \emph{Proceedings of the IEEE
  Conference on Computer Vision and Pattern Recognition (CVPR)}, June 2018.

\bibitem{32}
B.~AlBahar and J.-B. Huang, ``Guided image-to-image translation with
  bi-directional feature transformation,'' in \emph{Proceedings of the IEEE/CVF
  International Conference on Computer Vision (ICCV)}, October 2019.

\bibitem{33}
X.~Liang, C.~Xu, X.~Shen, J.~Yang, S.~Liu, J.~Tang, L.~Lin, and S.~Yan, ``Human
  parsing with contextualized convolutional neural network,'' in
  \emph{Proceedings of the IEEE International Conference on Computer Vision
  (ICCV)}, December 2015.

\bibitem{34}
Z.~Al-Halah, R.~Stiefelhagen, and K.~Grauman, ``Fashion forward: Forecasting
  visual style in fashion,'' in \emph{Proceedings of the IEEE International
  Conference on Computer Vision (ICCV)}, Oct 2017.

\bibitem{35}
C.~Yu, Y.~Hu, Y.~Chen, and B.~Zeng, ``Personalized fashion design,'' in
  \emph{Proceedings of the IEEE/CVF International Conference on Computer Vision
  (ICCV)}, October 2019.

\bibitem{36}
\BIBentryALTinterwordspacing
L.~Ma, X.~Jia, Q.~Sun, B.~Schiele, T.~Tuytelaars, and L.~V. Gool, ``Pose guided
  person image generation,'' \emph{CoRR}, vol. abs/1705.09368, 2017. [Online].
  Available: \url{http://arxiv.org/abs/1705.09368}
\BIBentrySTDinterwordspacing

\bibitem{37}
\BIBentryALTinterwordspacing
S.~E. Reed, Z.~Akata, X.~Yan, L.~Logeswaran, B.~Schiele, and H.~Lee,
  ``Generative adversarial text to image synthesis,'' \emph{CoRR}, vol.
  abs/1605.05396, 2016. [Online]. Available:
  \url{http://arxiv.org/abs/1605.05396}
\BIBentrySTDinterwordspacing

\bibitem{38}
\BIBentryALTinterwordspacing
Y.~Ma, J.~Jia, S.~Zhou, J.~Fu, Y.~Liu, and Z.~Tong, ``Towards better
  understanding the clothing fashion styles: A multimodal deep learning
  approach,'' in \emph{AAAI}, 2017, pp. 38--44. [Online]. Available:
  \url{http://aaai.org/ocs/index.php/AAAI/AAAI17/paper/view/14561}
\BIBentrySTDinterwordspacing

\bibitem{39}
K.~He, X.~Zhang, S.~Ren, and J.~Sun, ``Deep residual learning for image
  recognition,'' in \emph{Proceedings of the IEEE Conference on Computer Vision
  and Pattern Recognition (CVPR)}, June 2016.

\bibitem{40}
J.~Liu and H.~Lu, ``Deep fashion analysis with feature map upsampling and
  landmark-driven attention,'' in \emph{Proceedings of the European Conference
  on Computer Vision (ECCV) Workshops}, September 2018.

\bibitem{41}
\BIBentryALTinterwordspacing
Y.~Huang and T.~Huang, ``Outfit recommendation system based on deep learning,''
  in \emph{Proceedings of the 2nd International Conference on Computer
  Engineering, Information Science \& Application Technology (ICCIA
  2017)}.\hskip 1em plus 0.5em minus 0.4em\relax Atlantis Press, 2016/07, pp.
  164--168. [Online]. Available: \url{https://doi.org/10.2991/iccia-17.2017.26}
\BIBentrySTDinterwordspacing

\bibitem{42}
\BIBentryALTinterwordspacing
X.~Chen, Y.~Duan, R.~Houthooft, J.~Schulman, I.~Sutskever, and P.~Abbeel,
  ``Infogan: Interpretable representation learning by information maximizing
  generative adversarial nets,'' \emph{CoRR}, vol. abs/1606.03657, 2016.
  [Online]. Available: \url{http://arxiv.org/abs/1606.03657}
\BIBentrySTDinterwordspacing

\bibitem{43}
\BIBentryALTinterwordspacing
R.~de~Bem, A.~Ghosh, T.~Ajanthan, O.~Miksik, N.~Siddharth, and P.~H.~S. Torr,
  ``Dgpose: Disentangled semi-supervised deep generative models for human body
  analysis,'' \emph{CoRR}, vol. abs/1804.06364, 2018. [Online]. Available:
  \url{http://arxiv.org/abs/1804.06364}
\BIBentrySTDinterwordspacing

\bibitem{44}
\BIBentryALTinterwordspacing
K.~Qi, Q.~Guan, C.~Yang, F.~Peng, S.~Shen, and H.~Wu, ``Concentric circle
  pooling in deep convolutional networks for remote sensing scene
  classification,'' \emph{Remote Sensing}, vol.~10, no.~6, 2018. [Online].
  Available: \url{https://www.mdpi.com/2072-4292/10/6/934}
\BIBentrySTDinterwordspacing

\bibitem{45}
\BIBentryALTinterwordspacing
H.~Dong, X.~Liang, K.~Gong, H.~Lai, J.~Zhu, and J.~Yin, ``Soft-gated
  warping-gan for pose-guided person image synthesis,'' \emph{CoRR}, vol.
  abs/1810.11610, 2018. [Online]. Available:
  \url{http://arxiv.org/abs/1810.11610}
\BIBentrySTDinterwordspacing

\bibitem{46}
\BIBentryALTinterwordspacing
H.~Chui and A.~Rangarajan, ``A new algorithm for non-rigid point matching,'' in
  \emph{CVPR}, 2000, pp. 2044--2051. [Online]. Available:
  \url{https://doi.org/10.1109/CVPR.2000.854733}
\BIBentrySTDinterwordspacing

\bibitem{5}
Z.~Cao, T.~Simon, S.-E. Wei, and Y.~Sheikh, ``Realtime multi-person 2d pose
  estimation using part affinity fields,'' in \emph{Proceedings of the IEEE
  Conference on Computer Vision and Pattern Recognition (CVPR)}, July 2017.

\bibitem{6}
G.~Moon, J.~Y. Chang, and K.~M. Lee, ``Camera distance-aware top-down approach
  for 3d multi-person pose estimation from a single rgb image,'' in
  \emph{Proceedings of the IEEE/CVF International Conference on Computer Vision
  (ICCV)}, October 2019.

\bibitem{7}
G.~Huang, S.~Liu, L.~van~der Maaten, and K.~Q. Weinberger, ``Condensenet: An
  efficient densenet using learned group convolutions,'' in \emph{Proceedings
  of the IEEE Conference on Computer Vision and Pattern Recognition (CVPR)},
  June 2018.

\bibitem{8}
T.~Xiao, J.~Hong, and J.~Ma, ``Elegant: Exchanging latent encodings with gan
  for transferring multiple face attributes,'' in \emph{Proceedings of the
  European Conference on Computer Vision (ECCV)}, September 2018.

\bibitem{9}
N.~Jetchev and U.~Bergmann, ``The conditional analogy gan: Swapping fashion
  articles on people images,'' in \emph{Proceedings of the IEEE International
  Conference on Computer Vision (ICCV) Workshops}, Oct 2017.

\bibitem{11}
\BIBentryALTinterwordspacing
P.~Shen, X.~Lu, S.~Li, and H.~Kawai, ``Conditional generative adversarial nets
  classifier for spoken language identification,'' in \emph{Proc. Interspeech
  2017}, 2017, pp. 2814--2818. [Online]. Available:
  \url{http://dx.doi.org/10.21437/Interspeech.2017-553}
\BIBentrySTDinterwordspacing

\bibitem{12}
\BIBentryALTinterwordspacing
L.~Zhu, Y.~Chen, Y.~Lu, C.~Lin, and A.~L. Yuille, ``Max margin and/or graph
  learning for parsing the human body,'' in \emph{CVPR}, 2008. [Online].
  Available: \url{https://doi.org/10.1109/CVPR.2008.4587787}
\BIBentrySTDinterwordspacing

\bibitem{13}
K.~Gong, X.~Liang, D.~Zhang, X.~Shen, and L.~Lin, ``Look into person:
  Self-supervised structure-sensitive learning and a new benchmark for human
  parsing,'' in \emph{Proceedings of the IEEE Conference on Computer Vision and
  Pattern Recognition (CVPR)}, July 2017.

\bibitem{14}
R.~A. Güler, N.~Neverova, and I.~Kokkinos, ``Densepose: Dense human pose
  estimation in the wild,'' in \emph{Proceedings of the IEEE Conference on
  Computer Vision and Pattern Recognition (CVPR)}, June 2018.

\bibitem{15}
C.~Ledig, L.~Theis, F.~Huszar, J.~Caballero, A.~Cunningham, A.~Acosta,
  A.~Aitken, A.~Tejani, J.~Totz, Z.~Wang, and W.~Shi, ``Photo-realistic single
  image super-resolution using a generative adversarial network,'' in
  \emph{Proceedings of the IEEE Conference on Computer Vision and Pattern
  Recognition (CVPR)}, July 2017.

\bibitem{16}
W.~Shi, J.~Caballero, F.~Huszar, J.~Totz, A.~P. Aitken, R.~Bishop, D.~Rueckert,
  and Z.~Wang, ``Real-time single image and video super-resolution using an
  efficient sub-pixel convolutional neural network,'' in \emph{Proceedings of
  the IEEE Conference on Computer Vision and Pattern Recognition (CVPR)}, June
  2016.

\bibitem{17}
S.~Jandial, A.~Chopra, K.~Ayush, M.~Hemani, B.~Krishnamurthy, and A.~Halwai,
  ``Sievenet: A unified framework for robust image-based virtual try-on,'' in
  \emph{Proceedings of the IEEE/CVF Winter Conference on Applications of
  Computer Vision (WACV)}, March 2020.

\bibitem{18}
\BIBentryALTinterwordspacing
A.~Brock, J.~Donahue, and K.~Simonyan, ``Large scale {GAN} training for high
  fidelity natural image synthesis,'' in \emph{International Conference on
  Learning Representations}, 2019. [Online]. Available:
  \url{https://openreview.net/forum?id=B1xsqj09Fm}
\BIBentrySTDinterwordspacing

\end{thebibliography}

	\begin{IEEEbiography}[{\includegraphics[width=1in,height=1.25in,clip,keepaspectratio]{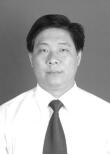}}]{Shanchen Pang }
		 (Member,IEEE)graduated from Tongji University in February 2008, majoring in computer software and theory. In March 2009, worked as a postdoctoral researcher at Tsinghua University.
		 
		 He is engaged in the (advanced) Petri net theory and application, formal methods, model checking, analysis and verification of the CSCW system, distributed concurrent systems in areas such as teaching and research work, presided over by the national natural science fund project, as academic backbone to participate in the national key basic research projects as a multiple international program committee member and chairman of the meeting, He is a reviewer of several important international and national journals.
	\end{IEEEbiography}

	\begin{IEEEbiography}[{\includegraphics[width=1in,height=1.25in,clip,keepaspectratio]{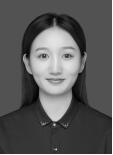}}]{Xixi Tao }
		received her bachelor's degree in network engineering from Inner Mongolia Agricultural University in 2019. She is currently studying for a master's degree at China University of Petroleum (East China).
		
		She is engaged in the research of generative adversarial networks in the field of deep learning and the theoretical research of explicable machine learning. It mainly includes online virtual fitting of 2D character pictures with generative adversarial network, extraction and control of abstract shape and texture features from network model.
	\end{IEEEbiography}
	
    \begin{IEEEbiography}[{\includegraphics[width=1in,height=1.25in,clip,keepaspectratio]{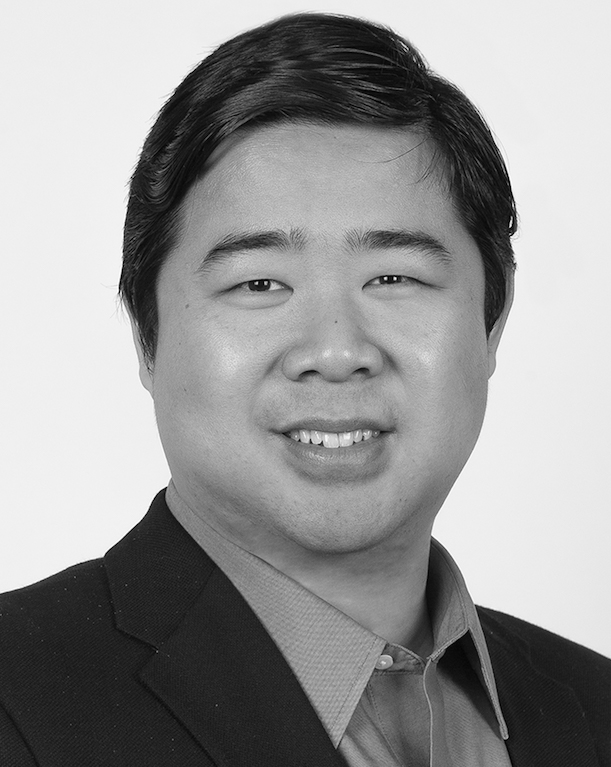}}]{Neal N. Xiong }
		(S’05–M’08–SM’12) received his both PhD degrees in Wuhan University (2007, about sensor system engineering), and Japan Advanced Institute of Science and Technology (2008, about dependable communication networks), respectively.
		
		He is current an Associate Professor (5rd year) at Department of Mathematics and Computer Science, Northeastern State University, OK, USA. And he has published over 200 international journal papers and over 100 international conference papers. His research interests include cloud computing, security and dependability, parallel and distributed computing, networks, and optimization theory.
		
		Dr. Xiong has been a General Chair, Program Chair, Publicity Chair, Program Committee member and Organizing Committee member of over 100 international conferences, and as a reviewer of about 100 international journals, including IEEE JSAC, IEEE SMC (Park: A/B/C), IEEE Transactions on Communications, IEEE Transactions on Mobile Computing, IEEE Trans. on Parallel and Distributed Systems.
	\end{IEEEbiography}

	\begin{IEEEbiography}[{\includegraphics[width=1in,height=1.25in,clip,keepaspectratio]{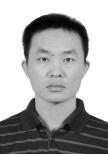}}]{Yukun Dong}
		graduated from the University of Petroleum (East China) in July 2003 with a major in communication engineering and received a master's degree in engineering from China University of Petroleum (East China) in July 2006. Now he is a teacher of software engineering department at China University of Petroleum (East China) and a doctoral candidate at Beijing University of Posts and Telecommunications. 
		
		He has published several academic papers. His current research interests include software testing and software quality measurement.
	\end{IEEEbiography}
\end{document}